\documentclass[conference,onecolumn]{IEEEtran}
\IEEEoverridecommandlockouts
\usepackage{cite}
\usepackage{amsmath,amssymb,amsfonts}
\usepackage{graphicx}
\usepackage[T1]{fontenc}
\usepackage[utf8]{inputenc}
\usepackage{textcomp}
\usepackage{xcolor}
\usepackage{booktabs}
\usepackage{url}
\def\BibTeX{{\rm B\kern-.05em{\sc i\kern-.025em b}\kern-.08em
    T\kern-.1667em\lower.7ex\hbox{E}\kern-.125emX}}
\begin{document}

\title{Evaluating Open-Weight LLMs for Turkish Domain Documents Under Retrieval and Hardware Constraints}

\author{%
\IEEEauthorblockN{
Imtiaz Ul Hassan\IEEEauthorrefmark{1},
\"Oyk\"u Akbulut\IEEEauthorrefmark{2},
Onur Kaya\IEEEauthorrefmark{2},
Ardhendu Behera\IEEEauthorrefmark{1},
Swagat Kumar\IEEEauthorrefmark{1},
Peter Matthew\IEEEauthorrefmark{1},
Yonghuai Liu\IEEEauthorrefmark{1}
}
\IEEEauthorblockA{
\IEEEauthorrefmark{1}\textit{Edge Hill University}, United Kingdom \\
hassani@edgehill.ac.uk,
beheraa@edgehill.ac.uk,
kumars@edgehill.ac.uk,
matthewp@edgehill.ac.uk,
liuyo@edgehill.ac.uk
}
\IEEEauthorblockA{
\IEEEauthorrefmark{2}\textit{Ermetal Otomotiv ve E\c{s}ya San. Tic. A.\c{S}.}, T\"urkiye \\
oyku.akbulut@ermetal.com,
onur.k@ermetal.com
}
}

\maketitle

\begin{abstract}
Most Turkish-capable large language models (LLMs) are evaluated using
general-purpose benchmarks rather than long, structurally complex domain
documents. This paper evaluates five open-weight 7B--8B models for Turkish
document question answering under a resource-constrained local deployment
setting. The primary benchmark contains 100 systematically validated questions
derived from a 109-page industrial R\&D report, and the evaluation protocol is
replicated using a second 112-page public-sector report and an independently
constructed 100-question set. All models are evaluated locally on an NVIDIA
RTX 3050 laptop GPU with 6\,GB VRAM using controlled prompting, decoding, and
4-bit quantisation.

The principal methodological contribution is an evidence-annotated evaluation
protocol that separates retrieval failure from downstream model reasoning
failure without requiring additional model calls. On the primary benchmark,
end-to-end accuracy ranges from 49\% to 75\%. Seven lexical, dense, and hybrid
retrieval configurations are additionally compared using 95\% Wilson intervals
and exact paired McNemar tests; none significantly outperforms the character
TF-IDF baseline on either document. Evidence recall saturates differently
across the two reports, showing that retrieval and effective context capacity
can be binding constraints for some documents but not others. These results
demonstrate that model selection, retrieval behaviour, and hardware limits must
be evaluated separately when deploying open-weight LLMs for Turkish domain
documents.
\end{abstract}

\begin{IEEEkeywords}
Large language models, Turkish NLP, document-grounded question answering,
retrieval-augmented generation, model evaluation, local deployment
\end{IEEEkeywords}

\section{Introduction}

The set of Turkish-capable open-weight LLMs has grown rapidly, with
Turkish-tuned models appearing alongside increasingly capable multilingual
systems~\cite{trendyolT1,kanarya,qwen25,llama3}. However, their effectiveness
on realistic Turkish document-analysis tasks remains unclear. Existing
evaluations rely largely on educational, general-knowledge, multiple-choice,
and curated reading-comprehension tasks~\cite{trmmlu,turkishmmlu,openllmtr,dogan,thquad}.
These settings generally do not evaluate long, structurally heterogeneous
Turkish documents processed through a retrieval pipeline, which is a common
deployment scenario for document-grounded LLM systems. We therefore ask whether current open-weight LLMs can reliably support
Turkish document-grounded question answering under realistic local
deployment constraints.

This question is particularly important for confidential industrial
documents, where cloud processing may conflict with organisational or
data-governance requirements. Such documents may originate in heterogeneous
formats but are commonly distributed as PDF reports containing text, tables,
and structural layout that must be recovered before retrieval. This setting
is relevant to the SynAM project, which combines data, machine learning, and
engineering research across academic and industrial partners. We therefore
study a Turkish industrial PDF using a locally deployed pipeline on a
consumer GPU.

Answering this question requires separating two failure modes that conventional
end-to-end evaluations often conflate. When a model answers a document question
incorrectly, either the model failed to reason over the supplied evidence, or
the retrieval system failed to provide the necessary evidence. Treating
pipeline accuracy as model accuracy silently charges the model for the
retriever's misses, even though the two failure modes require different
remedies. Recording, for each question, the specific evidence on which the
answer depends allows evidence recall to be measured independently of model
output and enables errors to be attributed either to the model or to the
retrieval apparatus. This decomposition becomes central to the analysis
presented in this paper.

We make three contributions. First, we construct and validate two
document-specific Turkish QA benchmarks, each containing 100 questions
covering factual retrieval, numerical reasoning, multi-step inference, and
abstention when the requested information is absent. The first benchmark is
based on a 109-page industrial R\&D report, while the second is based on a
112-page public-sector activity report. Each question records its supporting
evidence, enabling retrieval failures to be distinguished from model failures.

Second, we provide a controlled comparison of five open-weight 7B--8B models
under identical retrieval, prompting, decoding, quantisation, and hardware
conditions, using a question-specific grading procedure that accounts for
equivalent answer forms.

Third, we compare character and word TF-IDF, BM25, dense retrieval, and hybrid
retrieval under the same chunking and context-budget conditions. Evidence
recall is reported with 95\% Wilson intervals, and paired McNemar tests are
used to compare the retrieval methods. The results show that none of the
evaluated alternatives significantly outperforms character TF-IDF on either
document. They also show that the effect of limited context capacity differs
between the two documents.

\section{Related Work}
\label{sec:related}

\textbf{Turkish LLM evaluation.}
Existing Turkish LLM evaluations focus primarily on general knowledge,
educational content, and multiple-choice reasoning. MMLU~\cite{mmlu}
established the widely used multitask multiple-choice evaluation paradigm.
TurkishMMLU~\cite{turkishmmlu} applies this paradigm using more than 10,000
native Turkish questions drawn from the secondary-school curriculum, while
TR-MMLU~\cite{trmmlu} evaluates models using 6,200 Turkish multiple-choice
questions spanning educational and domain-specific categories. The OpenLLM
Turkish Leaderboard~\cite{openllmtr} aggregates evaluations based largely on
established general-purpose tasks, and Dogan et al.~\cite{dogan} compare
Turkish-capable models across several such benchmarks. THQuAD~\cite{thquad}
moves beyond multiple-choice evaluation by testing reading comprehension over
curated Turkish historical passages. These resources provide valuable evidence
about Turkish language understanding and general knowledge, but they do not
evaluate question answering over a long, structurally heterogeneous domain
document processed through a retrieval pipeline.

\textbf{Document-grounded QA and retrieval.}
Retrieval-augmented generation~\cite{rag} is a widely used architecture for
grounding language-model outputs in external documents. In such systems,
end-to-end accuracy depends not only on the generator but also on whether the
retriever supplies the evidence required to answer the question. Reporting
only final answer accuracy can therefore conflate retrieval failure with
generation or reasoning failure. The present work addresses this issue by
recording the supporting evidence required for each question and measuring
whether that evidence appears in the retrieved context.

This distinction is especially relevant under a fixed context window. Turkish
is morphologically rich, and its word forms may be segmented into relatively
large numbers of tokens by tokenisers developed primarily for multilingual or
English-dominant training corpora. In our deployment setting, this reduces the
amount of document text that can be supplied within the available context
budget. We therefore examine retrieval quality and effective context capacity
as components of the evaluation apparatus rather than treating them as
implementation details.

\textbf{Grading of generative QA.}
Open-ended QA cannot always be evaluated reliably by exact matching because
equivalent answers may differ in wording, inflection, numeric formatting, or
date representation. We therefore use question-specific grading rules for
numeric, entity, keyword, date, Boolean, multiple-choice, and abstention
answers, as detailed in Sec.~\ref{sec:grader}.

These limitations motivate an end-to-end evaluation beginning with the
original PDF and including extraction, retrieval, generation, and grading.

\section{Methodology}
\label{sec:method}

\subsection{Benchmark Design}

The main evaluation uses a 109-page Turkish industrial R\&D report in PDF
format containing running text, personnel and financial tables, publication
records, patents, collaborations, and project descriptions. The PDF is
converted to layout-preserving text before sectioning, chunking, and retrieval.
From this report, we construct and verify a 100-question benchmark.

To examine whether the evaluation procedure and the main observations depend
entirely on one document, we conduct a secondary cross-document validation
using the 112-page 2024 activity report of the Turkish Personal Data
Protection Authority~\cite{kvkk2024}. This public-sector report differs from the primary document in its
source, structure, and subject matter. A separate 100-question set is
constructed using the same question structure and evaluation procedure. This
second evaluation is used as a robustness check rather than as a comprehensive
multi-domain benchmark.

For each document, the questions are divided into four capability bands
comprising 30 Easy factual-retrieval questions, 30 Medium
numerical-reasoning questions, 20 Hard multi-step questions, and 20
Abstention questions whose answers are not present in the corresponding
report. Each question records its expected answer, supporting evidence, and
grading rule. The first three bands test increasingly demanding use of
retrieved evidence, while the Abstention band tests whether a model avoids
producing unsupported answers.

\begin{figure}[t]
\centering
\includegraphics[width=0.98\textwidth]{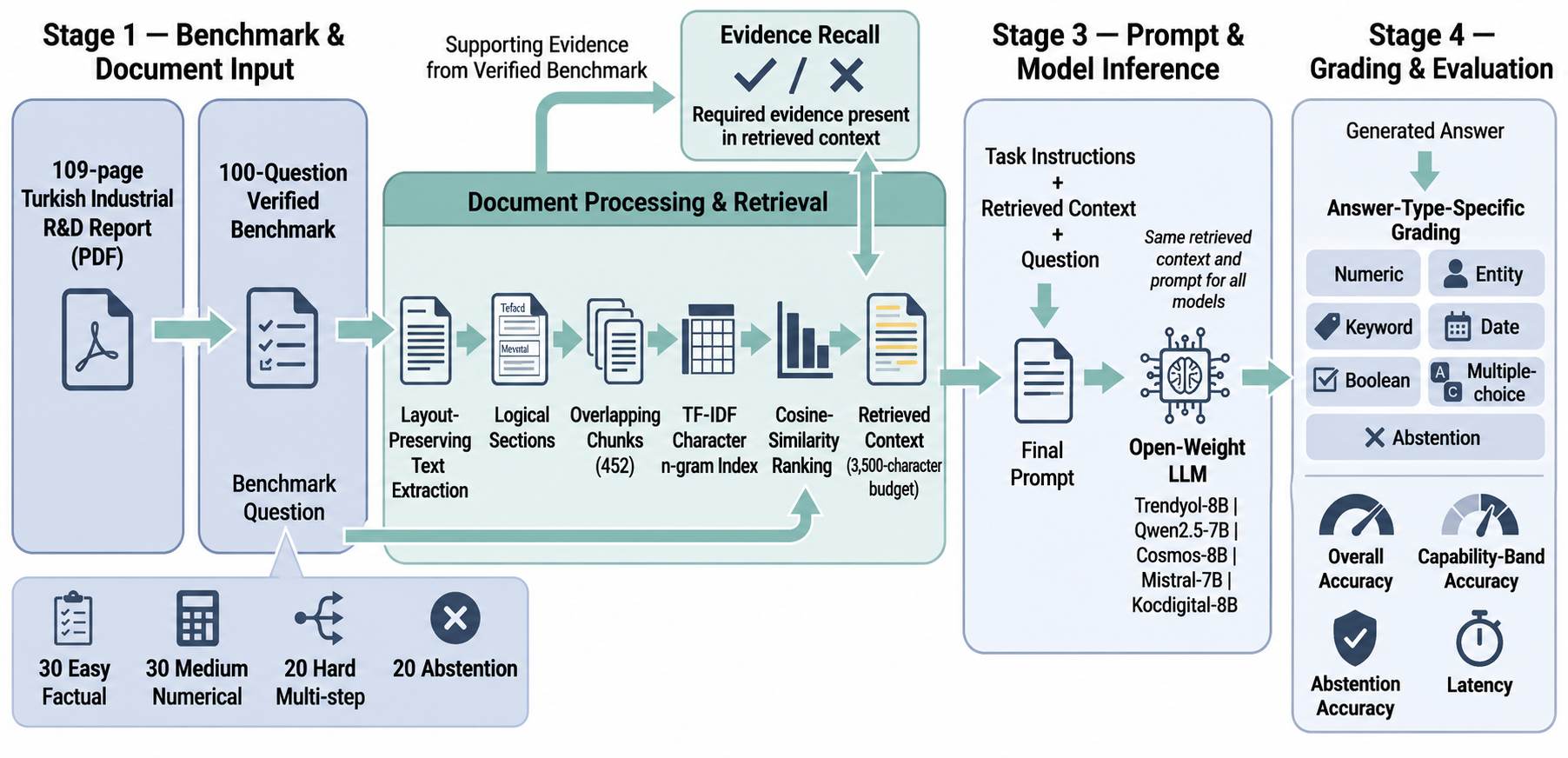}
\caption{End-to-end evaluation pipeline. The same retrieved context and prompt
are supplied to all models, while evidence recall is evaluated independently
from answer grading.}
\label{fig:pipeline}
\end{figure}

Figure~\ref{fig:pipeline} summarises the complete evaluation pipeline from
document preprocessing and question-aware retrieval to prompt construction,
model inference, evidence-recall analysis, and answer grading.

\subsection{Evaluation Protocol}

All models are evaluated under identical conditions. For a given question,
the retrieval stage is executed once and the resulting context is kept fixed
across all five models. Consequently, each model receives the same question,
retrieved evidence, prompt structure, decoding settings, and context budget.
This allows differences in answer quality to be compared without introducing
model-specific differences in retrieval.

Inference is performed locally on an NVIDIA RTX 3050 laptop GPU with
6\,GB VRAM. The 4,096-token context window was selected as a stable common
setting that fit within the available memory budget after accounting for
quantised model weights, KV-cache memory, prompt tokens, and generated output.
All models use temperature 0 and a fixed seed and are served locally via
Ollama.

We report overall accuracy, accuracy by capability band, abstention accuracy,
evidence recall, and latency. Overall accuracy is the proportion of the 100
questions answered correctly, while capability-band accuracy is calculated
within each question subset. Evidence recall is measured separately from
answer accuracy and indicates whether the evidence required to answer a
question is present in the retrieved context. This distinction allows
retrieval failures to be separated from failures that occur after the relevant
evidence has been supplied to the model.
The primary evaluation and the secondary cross-document validation use the
same model configurations, question distribution, prompting procedure,
decoding settings, and context budget. Unless otherwise stated, the model
comparisons use the character TF-IDF retriever described in
Sec.~\ref{sec:retrieval}. The retrieval comparison changes only the ranking
method while keeping chunking, context selection, and the character budget
fixed.

Evidence recall is calculated only for questions whose required supporting
figures are explicitly recorded and can be checked in the retrieved context.
Abstention questions are excluded because the requested information is absent
by construction. Questions whose expected answer is a Boolean value or a
multiple-choice label are also excluded when that label does not represent
the supporting evidence in the document.

Uncertainty in evidence-recall estimates is reported using 95\% Wilson score
intervals. Retrieval methods are compared using exact paired McNemar tests
because each method is evaluated on the same questions. A difference is
treated as statistically significant when \(p < 0.05\).
\subsection{Evaluated Models}
\label{sec:models}

We evaluate five open-weight models in the 7B--8B parameter range:
Trendyol-LLM-8B-T1, Qwen2.5-7B, Cosmos Turkish-Llama-8B-Instruct,
Mistral-7B, and Kocdigital-LLM-8B. Trendyol-LLM-8B-T1 is a
Turkish--English reasoning model built on Qwen3-8B~\cite{trendyolT1}.
Qwen2.5-7B and Mistral-7B serve as general multilingual
baselines~\cite{qwen25,mistral7b}. Cosmos and Kocdigital are
Turkish-adapted Llama-based models~\cite{cosmos,kesgin2024optimizing,kocdigital}.

The models were selected because they have comparable parameter counts,
represent both Turkish-specialised and general multilingual systems, and can
be evaluated using the same Q4\_K\_M quantisation and 6\,GB GPU constraint.
Their parameter counts are reported in Table~\ref{tab:main}, while the cited
model cards provide the corresponding model configurations.

\subsection{Document Processing, Retrieval, and Prompt Construction}
\label{sec:retrieval}

As shown in Fig.~\ref{fig:pipeline}, each source PDF is converted to text using
layout-preserving extraction so that line structure and the relative
arrangement of table content are retained as far as possible. The extracted
text is organised into logical sections and divided into overlapping chunks.
This produces 452 chunks for the primary industrial report and 348 chunks for
the secondary public-sector report.

The main model evaluation uses TF-IDF character \(n\)-gram features. For each
benchmark question, the question text is represented in the same feature space
as the document chunks. Cosine similarity is calculated between the question
and each chunk, producing a ranked list of candidate evidence. Chunks are
selected in descending similarity order until the 3,500-character context
budget is reached.

For the retrieval comparison, seven configurations are evaluated under the
same chunking and context-selection conditions. These comprise character
TF-IDF, word TF-IDF, BM25, multilingual E5-small, multilingual E5-base,
reciprocal-rank fusion of character TF-IDF and BM25, and reciprocal-rank
fusion of character TF-IDF and E5-base. Only the ranking method is changed.
The chunk boundaries, context budget, overlap handling, and document-order
assembly remain fixed.

The selected chunks form the evidence context for the question. The final
prompt consists of task instructions, the retrieved context, and the question
itself. The instructions require the model to answer using the supplied
document evidence and to abstain when the requested information cannot be
supported. For a given document and question, the same assembled prompt is
supplied to each of the five models.

\subsection{Grading and Score Calculation}
\label{sec:grader}

Exact string matching is unsuitable for model-generated answers because
semantically equivalent Turkish responses may differ in number formatting,
date representation, inflection, wording, or explanatory text. Each benchmark
question is therefore assigned an answer-type-specific grading rule before
model evaluation. The supported answer types include numeric, multiple-choice,
Boolean, entity, keyword, date, and abstention responses.

The grading rule defines what constitutes a correct response for each
question. Numeric questions are judged using the required numerical value
rather than the surrounding explanation. Years and exact integer counts
require an exact value, while a small numerical tolerance is retained for
derived ratios, percentages, and rounded decimal values. Entity and keyword
questions are checked against predefined acceptable forms. Multiple-choice
and Boolean questions are mapped to their expected category, and Abstention
questions are scored as correct only when the model indicates that the
requested information cannot be established from the supplied evidence. Each
question receives a binary score of 1 for a correct answer and 0 otherwise.

Overall accuracy is the proportion of correctly answered questions, with the
same calculation applied within each capability band. Evidence recall is
computed independently from the expected answer string. A question is counted
as recalled only when all recorded source figures required to produce its
answer are present in the retrieved context. This is particularly important
for numerical-reasoning questions whose final answers are calculated and may
not appear literally in the source document.

The reference answers and supporting evidence were checked against the cited
pages of the corresponding source document. Derived answers were recalculated
from the source tables, while the targets of Abstention questions were checked
using full-document search. The configured grading rules were also tested
using the expected answers before model outputs were evaluated.

\section{Results}
\label{sec:results}

\begin{figure}[t]
\centering
{\includegraphics[width=\textwidth]{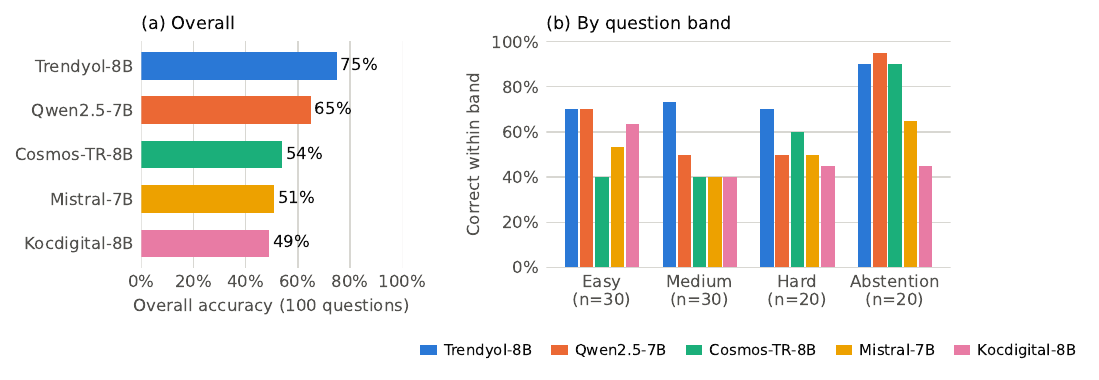}}
\caption{Model performance. (a) Overall accuracy on the 100-question
benchmark. (b) Accuracy by capability band. Trendyol leads overall, with its
largest advantage in the Medium band, while abstention performance reveals a
different ordering from factual accuracy.}
\label{fig:results}
\end{figure}

\subsection{Accuracy and Inference Cost}

Fig.~\ref{fig:results} and Table~\ref{tab:main} show that overall accuracy
ranges from 49\% to 75\%, a 26-point spread despite comparable model sizes and
identical evaluation conditions. Table~\ref{tab:cost} reports latency measured
on the fixed 20-question timing sample.

\begin{table}[htbp]
\centering

\begin{minipage}[t]{0.49\textwidth}
\vspace{0pt}
\centering
\caption{Benchmark accuracy by capability band.}
\label{tab:main}

\resizebox{\linewidth}{!}{%
\begin{tabular}{lcccccc}
\toprule
\textbf{Model} & \textbf{Par.} & \textbf{Acc.} &
\textbf{E} & \textbf{M} & \textbf{H} & \textbf{Abs} \\
 & & & /30 & /30 & /20 & /20 \\
\midrule
Trendyol-8B   & 8.2B & \textbf{75\%} & 21 & 22 & 14 & 18 \\
Qwen2.5-7B    & 7.6B & 65\%          & 21 & 15 & 10 & \textbf{19} \\
Cosmos-8B     & 8.0B & 54\%          & 12 & 12 & 12 & 18 \\
Mistral-7B    & 7.2B & 51\%          & 16 & 12 & 10 & 13 \\
Kocdigital-8B & 8.0B & 49\%          & 19 & 12 &  9 &  9 \\
\bottomrule
\end{tabular}%
}
\end{minipage}
\hfill
\begin{minipage}[t]{0.49\textwidth}
\vspace{0pt}
\centering
\caption{Inference latency on the timing sample.}
\label{tab:cost}

\resizebox{\linewidth}{!}{%
\begin{tabular}{lcccc}
\toprule
\textbf{Model} & \textbf{Mean} & \textbf{Median} &
\textbf{Est. 100 q} & \textbf{s/correct} \\
 & (s/q) & (s/q) & (min) & \\
\midrule
Trendyol-8B   & 23.86 & 23.36 & 39.8 & 31.8 \\
Qwen2.5-7B    &  1.69 &  1.47 &  2.8 & \textbf{2.6} \\
Cosmos-8B     &  3.21 &  3.21 &  5.4 &  6.0 \\
Mistral-7B    &  2.63 &  2.42 &  4.4 &  5.2 \\
Kocdigital-8B &  5.75 &  3.30 &  9.6 & 11.7 \\
\bottomrule
\end{tabular}%
}
\end{minipage}

\end{table}

Trendyol and Qwen2.5 tie on Easy questions at 21/30, but Trendyol reaches
22/30 on Medium questions compared with 15/30 for Qwen2.5 and 12/30 for the remaining models. This pattern suggests that reasoning over retrieved figures, rather than extraction alone, contributes to Trendyol's advantage. The multilingual Qwen2.5 model outperforms Cosmos and Kocdigital, showing that Turkish-specific adaptation alone does not guarantee stronger
document-grounded performance.

The secondary cross-document evaluation produced overall accuracies
of 82\%, 65\%, 60\%, 58\%, and 51\% for Trendyol, Qwen2.5, Cosmos,
Kocdigital, and Mistral, respectively. The three strongest models
retained their ordering from the primary benchmark, while the two
lowest-ranked models exchanged positions.

\subsection{Faithfulness}

Abstention performance differs from overall accuracy. Qwen2.5 achieves the
highest abstention score at 19/20, while Cosmos and Trendyol each reach 18/20.
Kocdigital correctly abstains on only 9/20 questions, despite scoring 19/30 on
Easy questions. This makes it less suitable for factual report-search settings
where unsupported answers are costly.

\begin{figure}[t]
\centering
\includegraphics[height=4.6cm,keepaspectratio]{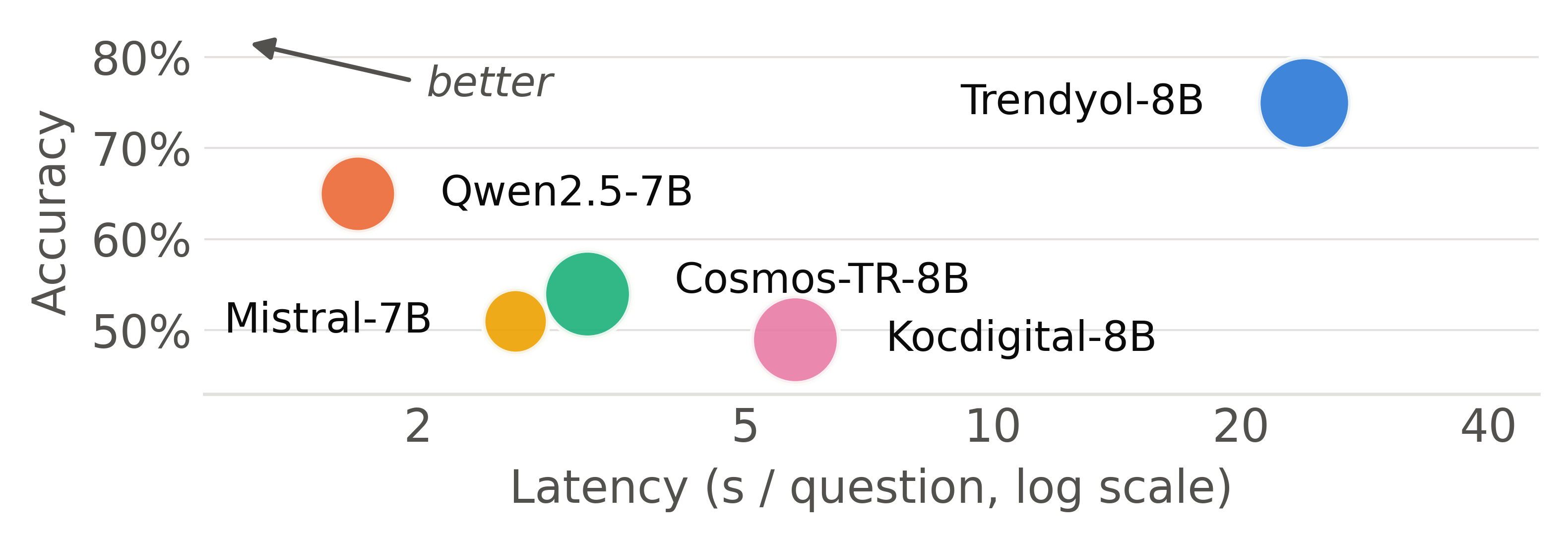}
\caption{Accuracy--latency trade-off on the target 6\,GB GPU. The upper-left
region is preferable.}
\label{fig:tradeoff}
\end{figure}

Qwen2.5 provides the strongest efficiency trade-off, achieving the
second-highest accuracy with the lowest mean latency. Trendyol gains ten
accuracy points but is approximately fourteen times slower per question.
Its final answers contain 207 characters on average after scratchpad removal,
compared with 44 for Qwen2.5. Trendyol-T1 also inherits Qwen3's thinking mode,
which generates intermediate reasoning tokens before the final answer. These
tokens are removed before grading and output-length measurement but still
contribute to inference latency. Extraction, indexing, retrieval, and grading
contribute negligible runtime compared with model inference.

\subsection{Retrieval Comparison}

Seven retrieval configurations were compared using the same chunking and
3,500-character context budget. On the primary report, evidence recall was
57\% for character TF-IDF, 65\% for BM25, 59\% for E5-base, and 63\% for the
strongest hybrid. On the secondary report, the corresponding values were
91\%, 86\%, 90\%, and 91\%. The character TF-IDF estimates had 95\% Wilson
intervals of [44, 70] and [82, 96], respectively. Exact paired McNemar tests
found no significant improvement from any lexical, dense, or hybrid
alternative on either document, with all \(p > 0.05\). The results therefore
show no consistent advantage for a more complex retriever under the evaluated
conditions.

\section{Error Decomposition}
\label{sec:analysis}

End-to-end accuracy includes both model and retrieval failures. We separate
these cases by checking whether all source figures required for an answer are
present in the retrieved context.

At the 3,500-character operating budget, evidence recall is 31/54, or 57\%,
on the primary report and 70/77, or 91\%, on the secondary report. Primary
recall is 74\% for Easy, 43\% for Medium, and 58\% for Hard questions. The
corresponding secondary results are 96\%, 93\%, and 79\%. Only questions with
explicitly recorded source figures are included. Abstention questions and
Boolean or multiple-choice labels without retrievable source figures are
excluded. Differences in evidence annotation mean that the two overall values
should not be interpreted as a controlled ranking of the documents.

Diagnostic budgets of 7,000 and 9,000 characters increase primary recall to
72\% and 76\%, while secondary recall changes only to 94\% and 95\%. These
larger budgets contain the evidence selected at smaller budgets and therefore
indicate where recall saturates. They are not used for model evaluation because
a 7,000-character context plus the generation allowance exceeds the common
4,096-token limit. Retrieval is therefore a substantial constraint for the
primary report but is less restrictive for the secondary report.

\section{Conclusion}
\label{sec:conclusion}
We evaluated five open-weight 7B--8B models for Turkish document QA
under a fixed retrieval pipeline and a 6\,GB consumer-GPU constraint.
On the primary industrial report, accuracy ranged from 49\% to 75\%,
with Trendyol strongest overall and Qwen2.5 providing the best
accuracy--latency trade-off. On the secondary public-sector report,
the corresponding model accuracies ranged from 51\% to 82\%, and the
three strongest models retained their ordering.

The evidence-based analysis showed 57\% retrieval recall on the primary report
and 91\% on the secondary report at the same operating budget. None of the
evaluated lexical, dense, or hybrid retrieval alternatives produced a
significant improvement over character TF-IDF. These results show that the
effect of retrieval and context limitations depends on the source document.

The primary benchmark remains based on one industrial R\&D report,
while the public-sector report serves as a cross-document robustness
check. Replication using legal, medical, and other sector-specific
documents is needed before broader conclusions can be drawn.

\section*{Acknowledgment}

This work was partially supported by the European Union's Horizon Europe
research and innovation programme under the Marie Sk\l{}odowska-Curie
Actions, Grant Agreement No.~101129996 (SynAM), and by UK Research and
Innovation through the Engineering and Physical Sciences Research Council
under the UK government's Horizon Europe funding guarantee, Grant
No.~EP/Y036778/1.

\end{document}